\documentclass[letterpaper, 10 pt, conference]{ieeeconf}  

\IEEEoverridecommandlockouts                              
\usepackage{todonotes}
\usepackage{graphicx} 
\usepackage{times} 
\usepackage{amsmath} 
\usepackage{amssymb}  
\usepackage{todonotes}
\usepackage{verbatim} 
\usepackage{cite}
\usepackage{subcaption}
\usepackage[hyphens]{url}
\usepackage[ruled,vlined]{algorithm2e}
\usepackage{float}
\usepackage{tabulary}
\usepackage{tabularx}
\usepackage{booktabs} 
\usepackage{fixltx2e}
\usepackage{dblfloatfix}
\usepackage{lipsum,amsmath,multicol}
\usepackage{tensor}
\usepackage{multirow}
\usepackage{balance}

\def\fps@figure{htp}
\def\fps@table{htp}

\newcommand{\bi}{\begin{itemize}}
\newcommand{\ei}{\end{itemize}}

\newcommand{\bfig}{\begin{figure}}
\newcommand{\efig}{\end{figure}}

\newcommand{\benum}{\begin{enumerate}}
\newcommand{\eenum}{\end{enumerate}}

\newcommand{\be}{\begin{equation}}
\newcommand{\ee}{\end{equation}}

\newcommand{\ba}{\begin{eqnarray}}
\newcommand{\ea}{\end{eqnarray}}

\newcommand{\unit}[1]{\mbox{$\rm \,#1$}}

\usepackage{color}

\definecolor{CommentRed}{rgb}{0.7,0,0}
\definecolor{CommentBlue}{rgb}{0,0,0.7}
\definecolor{CommentDG}{rgb}{0,0.6,0}

\newcommand{\footnoteref}[1]{\textsuperscript{\ref{#1}}}

\makeatletter
\newenvironment{myalign*}{%
  \setlength{\mathindent}{0pt}%
  \setlength{\abovedisplayskip}{-\baselineskip}%
  \setlength{\abovedisplayshortskip}{\abovedisplayskip}%
  \start@align\@ne\st@rredtrue\m@ne
}%
{\endalign}
\makeatother

\title{\LARGE \bf
In-Field Peduncle Detection of Sweet Peppers for Robotic Harvesting: a comparative study
}

\author{$\text{Chris Lehnert}^*$, $\text{Chris McCool}^{*}$ and $\text{Tristan Perez}^{*}$ \thanks{* Science and Engineering Faculty, Queensland University of Technology, Brisbane, Australia.
 \texttt{\{c.lehnert, c.mccool, tristan.perez\}@qut.edu.au}}
}

\begin{document}

\maketitle

\thispagestyle{empty}
\pagestyle{empty}


\begin{abstract}
Robotic harvesting of crops has the potential to disrupt current agricultural practices.
A key element to enabling robotic harvesting is to safely remove the crop from the plant which often involves locating and cutting the peduncle, the part of the crop that attaches it to the main stem of the plant.

In this paper we present a comparative study of two methods for performing peduncle detection.
The first method is based on classic colour and geometric features obtained from the scene with a support vector machine classifier, referred to as \textit{PFH-SVM}.
The second method is an efficient deep neural network approach, \textit{MiniInception}, that is able to be deployed on a robotic platform.
In both cases we employ a secondary filtering process that enforces reasonable assumptions about the crop structure, such as the proximity of the peduncle to the crop.
Our tests are conducted on Harvey, a sweet pepper harvesting robot, and is evaluated in a greenhouse using two varieties of sweet pepper, \textit{Ducati} and \textit{Mercuno}.
We demonstrate that the \textit{MiniInception} method achieves impressive accuracy and considerably outperforms the \textit{PFH-SVM} approach achieving an $F_{1}$ score of 0.564 and 0.302 respectively.


\end{abstract}

\section{INTRODUCTION}
\label{sec:intro}

Robotics in horticulture will play a key role in improving productivity and increasing crop quality~\cite{kondo2011agricultural}.
In particular, autonomously harvesting high value crops has the potential to overcome labour shortages and provide greater consistency in crop quality.
Achieving this is extremely challenging due to the difficulties in manipulating the crop and also for the vision systems that enable the robot to understand the scene.

\begin{figure}
	\centering
	\begin{subfigure}[b]{\columnwidth}\centering
		\includegraphics[width=0.8\columnwidth]{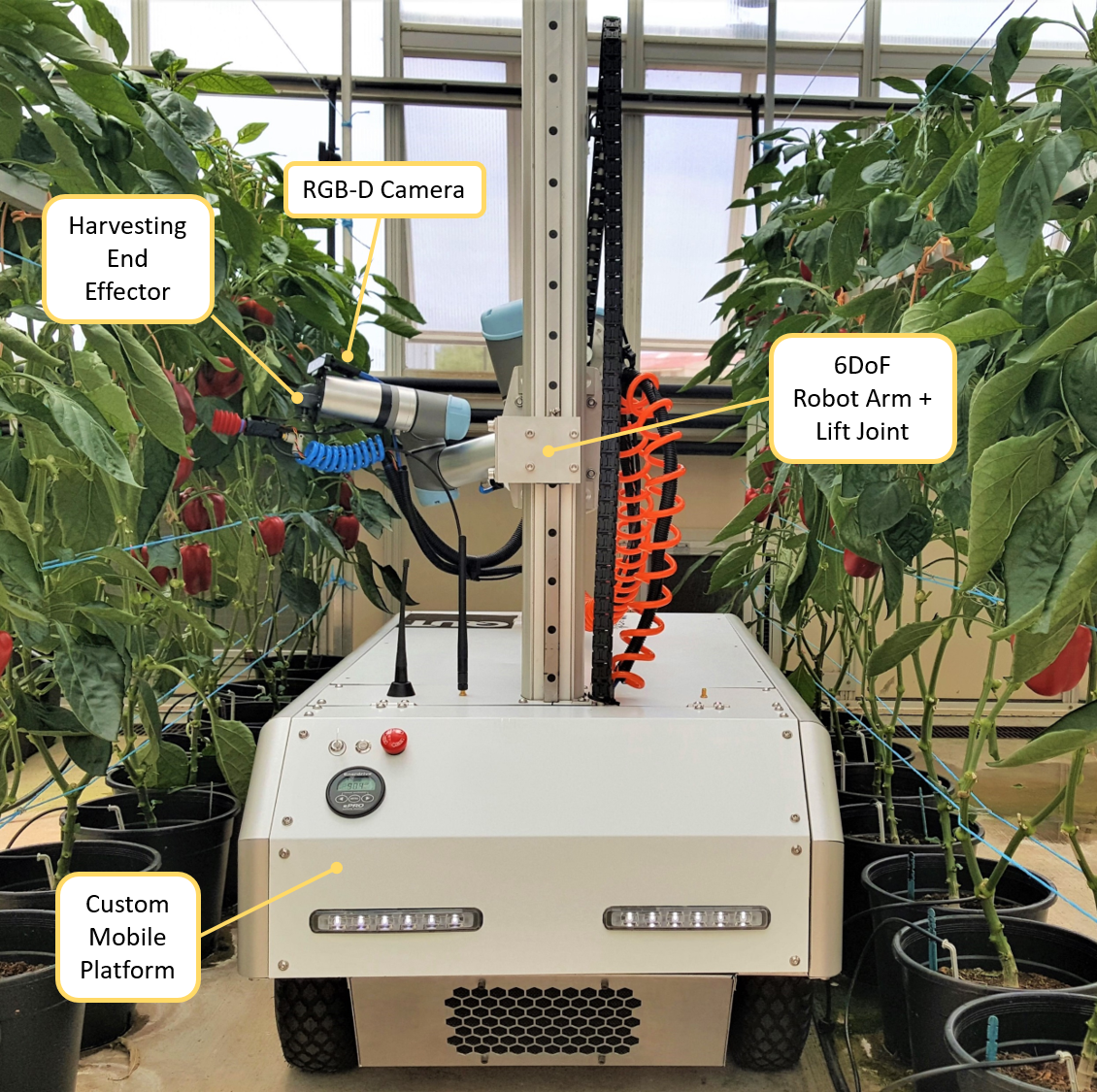}
		 \caption{}
	\end{subfigure}
    \\
	\begin{subfigure}[b]{\columnwidth}\centering
    	\includegraphics[width=0.8\columnwidth]{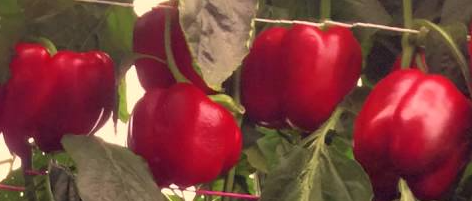}
    	\caption{}
    \end{subfigure}

		\caption{Top (a) is an example image of QUT's Harvey, an autonomous sweet pepper harvester prototype within a field testing greenhouse facility. 
		Bottom (b) is an example image of the sweet pepper crop and the challenging positions and orientations that the peduncle can present with.}
	\label{fig:harvey_pic}
	\label{fig:first_image}
\vspace{-20pt}
\end{figure}

Considerable progress has recently been made towards autonomous harvesting.
Bac et al.~\cite{Bac17_1} presented a robotic sweet pepper harvester with two different end effectors and by applying simplifications to the crop, such as removing leaves, they were able to autonomously harvest up to 33\% of the crop.
Lehnert et al.~\cite{lehnert2017autonomous, Lehnertb} also developed a robotic sweet pepper harvester, see Figure~\ref{fig:harvey_pic} (a), which had a multi-modal end effector.
This end effector harvests the crop by first attaching to it using a suction cup and then cutting the peduncle using a cutting tool.
After applying simplifications to the crop, Harvey demonstrated a success rate of 58\%.

One of the critical elements for these two harvesting robots is the accurate localisation of the peduncle, the part of the crop that attaches it to the main stem of the plant.
Accurately locating the peduncle could allow for improved rates of crop detachment as well as making the process safer for the plant.
Additionally, retaining the proper peduncle maximises the storage life and market value of crops such as sweet pepper.

Peduncle detection is challenging due to the varying lighting conditions and the presence of occlusions by leaves or other crops as shown in Fig.~\ref{fig:first_image} (b). 
Further confounding the issue is the fact that the peduncle for sweet peppers are green as are the other parts of the plant, including the crop; the crop can vary in colour from green through to red (with other variations possible).
Recently, Sa et al.~\cite{Sa17_1} proposed a peduncle detection system to address this challenge based on classic colour and geometry features (such as point feature histograms~\cite{Paulus2013}) used with a support vector machine (SVM), referred to as \textit{PFH-SVM}.
At a similar time, an efficient deep learning method was proposed by McCool et al.~\cite{McCool17_1} for weed detection which achieved considerably better performance than traditional features, referred to as \textit{MiniInception}.

Inspired by these two recent advances for robotic vision, we explore how to deploy an accurate peduncle detection system on a robotic harvester and provide a comparative study.
We concentrate on the task of sweet pepper peduncle detection which is deployed onto Harvey~\cite{lehnert2017autonomous} a sweet pepper robotic harvesting platform, shown in Fig.~\ref{fig:first_image}.
In particular, we compare and contrast the \textit{PFH-SVM} approach which is based on classic features against the \textit{MiniInception} approach which is an efficient deep neural network method. 
For both of these approaches we employ a secondary filtering process that enforces additional structural constraints based on reasonable assumptions about the crop structure.
Our tests are conducted on a autonomous sweet pepper harvesting robot, Harvey, and is deployed in a greenhouse using two varieties of sweet pepper, \textit{Ducati} and \textit{Mercuno}.
We demonstrate that the \textit{MiniInception} method considerably outperforms the \textit{PFH-SVM} approach achieving an $F_{1}$ score of 0.564 and 0.302 respectively.

The remainder of this paper is structured as follows.
Section~\ref{sec:background} introduces related work and background. 
Section~\ref{sec:detection_algorithm} describes sweet pepper peduncle detection and we present our experimental results in Section~\ref{sec:results}. 
Conclusions are drawn in Section~\ref{sec:conclusion}.




\section{Related Work/Background}\label{sec:background}

Peduncle detection has been primarily considered in highly controlled scenarios.
These controlled scenarios often consist of an inspection chamber with ideal lighting, no occlusion and high-quality static imagery.
Cubero et al.~\cite{Cubero201462} presented a peduncle detection system based on the radius and curvature signatures of fruits.
This was successfully applied to mulitple crops and made use of the Euclidean distance along with the angular rate of change between points on the contour.
Blasco et al.~\cite{blasco2003machine,Ruiz:1996aa} proposed an approach to detect peduncles for oranges, peaches, and apples based on a Bayesian discriminant model of RGB colour information.
This provided a set of colour segmented areas whose area was used as the feature to assign the class label.
The methods of Cubero et al. and Blasco et al. were deployed in highly controled settings and able to achieve between 95\% and 100\% accuracy.

Peduncle detection for in-field crop, outside of highly controlled scenarios, has recently received greater attention.
Hyper-spectral images were used by Bac et al.~\cite{Bac2013a} to classify different parts of sweet pepper plants. 
Despite using hyper-spectral images they achieved an accuracy of just of 40\% for peduncles and similar rates for other parts of the plant.
A laser range finder was used by Paulus et al.~\cite{Paulus2013} to differentiate between structures such as the leaf and stem using a point feature histogram achieving impressive accuracy.
Peduncle detection for strawberries was explored by Hayashi et al.~\cite{Hayashi:2010aa} using colour imagery. 
Using prior knowledge a region of interest (ROI) was defined and the boundary point between a fruit (which was red) and peduncle (green) was detected using colour information. 

More recently, Sa et al.~\cite{Sa17_1} proposed the use of colour and geometry features to accurately detect the peduncles of sweet pepper.
The colour features were first converted from the RGB space to the more robust HSV space and appended to a geometry feature vector consisting of a point feature histogram, similar to the work of Paulus et al.
Impressive results were achieved for detecting the peduncle of green, mixed colour (green and red) and red sweet pepper. 
However, a downside to the approach of Sa et al. is that it requires the accurate 3D reconstruction of the scene and its generalisation and performance in the field has yet to be demonstrated.



\section{Methods}\label{sec:detection_algorithm}

\begin{figure}[!t]
\centering

	\begin{subfigure}[thb]{0.45\columnwidth}
		\includegraphics[width=\textwidth]{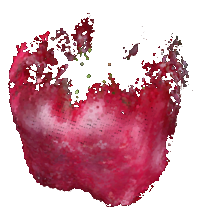}
    \end{subfigure}	
	\begin{subfigure}[thb]{0.45\columnwidth}
		\includegraphics[width=\textwidth]{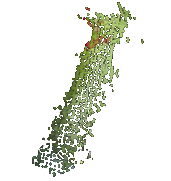}
	\end{subfigure} 

	\caption{Example 3D point cloud of manually annotated ground truth for sweet peppers (left) and peduncles (right).}
	\label{fig:dataset_instances}
\vspace{-10pt}
\end{figure}

Accurately detecting the peduncle of a sweet pepper is highly desirable as it enables robotic harvesters, such as Harvey (see Fig.~\ref{fig:first_image} (a)), to successfully cut the crop without damaging the surrounding plant (main stem and branches). 
Peduncle detection for fruit, and in particular sweet peppers, is challenging as the peduncles are highly curved and sometimes even flattened against the top of the fruit as shown in Fig.~\ref{fig:first_image} (b).

In this paper we explore the use of two approaches for peduncle detection.
The first approach we consider is the peduncle detection approach of Sa et al.~\cite{Sa17_1} which achieved impressive results for sweet pepper peduncle detection.
However, its downside is that it takes as input a dense reconstruction of the scene using Kinect Fusion. 
Also, this approach has yet to be deployed to a robotic system.

The second approach is based on the lightweight deep convolutional neural networks proposed by McCool et al.~\cite{McCool17_1}.
These networks allow for the tradeoff of accuracy and complexity to enable training of a DCNN, with limited data, which can be deployed in agricultural settings.
Initial results showed its potential for weed segmentation, however, its application to other fields remained untested.

For both approaches we apply a post filtering stage.
This allows us to remove incorrect points using physical characteristics of the peduncle and is used to cluster the final points to produce the candidate (or most likely) peduncle position.
Before we detail the two algorithms and the filtering stage, we first describe the robotic platform on which they are deployed, Harvey the sweet pepper harvester.

\begin{figure*}[t]
\begin{center}
\includegraphics[width=0.8\textwidth]{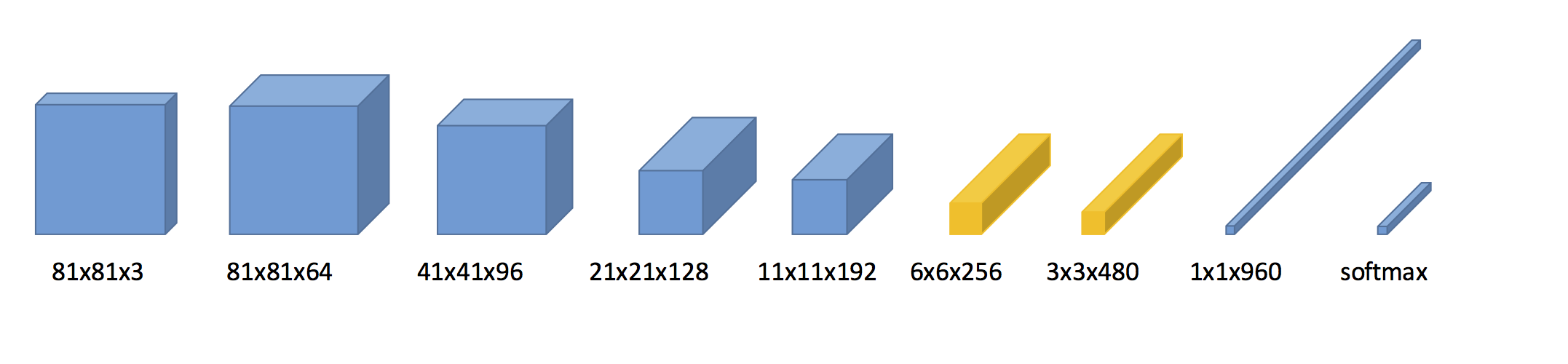}
\end{center}
\vspace{-10pt}
	\caption{An illustration of the \textit{MiniInception} architecture which consists of 8  convolutional  layers  and one fully connected layer leading to 5.1M parameters. The two layers highlighted in yellow make use of Inception modules~\cite{Szegedy15_1:conference}.}
	\label{fig:miniincetion_arch}
\vspace{-15pt}
\end{figure*}


\begin{figure*}[b]
	\centering
	\includegraphics[width=\textwidth]{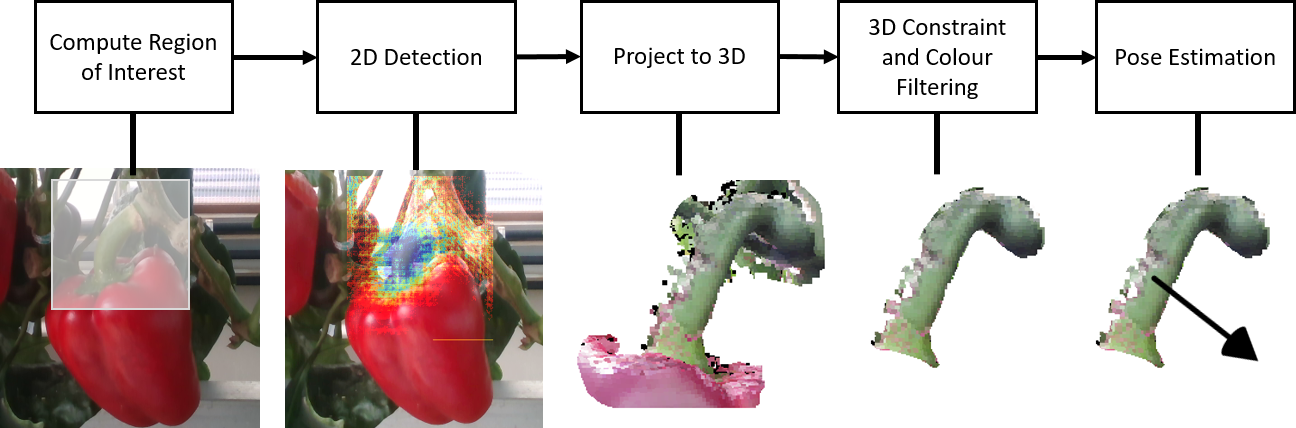}
	\vspace{-15pt}
	\caption{Steps for CNN peduncle detection. Firstly a region of interest is computed given prior detected sweet pepper points. The ROI masked image is then passed into the CNN which returns detected peduncle points. The detected points are then projected to Euclidean space using the corresponding depth image from the RGD-D camera. The projected points are then filtered using 3D constraints, colour information and cluster size. The filtered peduncle points are then used to estimate a cutting pose using aligned with the centroid of the points.   }
	\label{fig:pipe_line}
\end{figure*}

\subsection{Harvey: autonomous harvesting system}

We have developed a prototype robotic sweet pepper harvester, ``Harvey", which has 7 Degrees of Freedom (DoF), including a 6 DoF robotic arm and a 1 DoF linear lift axis on a custom mobile base platform as shown in Fig.~\ref{fig:harvey_pic} (a). 
The robot arm has a customised harvesting tool mounted on its end effector that both grips each pepper with a suction cup, and cuts the peduncle with an oscillating cutting blade. 
An Intel i7 PC records incoming RGB-D data (640$\times$480) from a eye-in-hand colour-depth camera. 

In~\cite{lehnert2017autonomous}, Harvey picks sweet peppers by first taking an image of the scene at a wider field of view to obtain an initial estimate of sweet peppers in the scene. 
Harvey decides which sweet pepper to harvest next and moves the eye-in-hand camera to a close up view of the sweet pepper. 
The sweet pepper is then detected within a colour point cloud. 
Grasp points are then selected using the surface properties of the sweet pepper. 
In work presented by Lehnert et al. the peduncle is assumed to be located above the centre of the top face of the detected sweet pepper and the cutting tool is then used to separate the sweet pepper from the plant after a successful attachment. 

The performance of the robotic harvester relies on both an attachment success and cutting success. 
In its current implementation, the cutting stage assumes the location of the peduncle is at the centre of the top face of a sweet pepper and extends directly vertical to this surface. 
This is prone to errors if the peduncle is misshapen or abnormal. 
Therefore, by accurately detecting the location and geometry of the peduncle will enable the robotic harvester to improve the success rate of cutting sweet pepper peduncles.
Despite these issues, using the above assumption led to an attachment and detachment rate of up to 58\%.

Below we describe the two systems that we analyse in order to improve the peduncle detection rate and in so doing improve the overall harvesting rate of Harvey.


\begin{figure*}[tb]
\begin{center}
\includegraphics[width=0.3\textwidth]{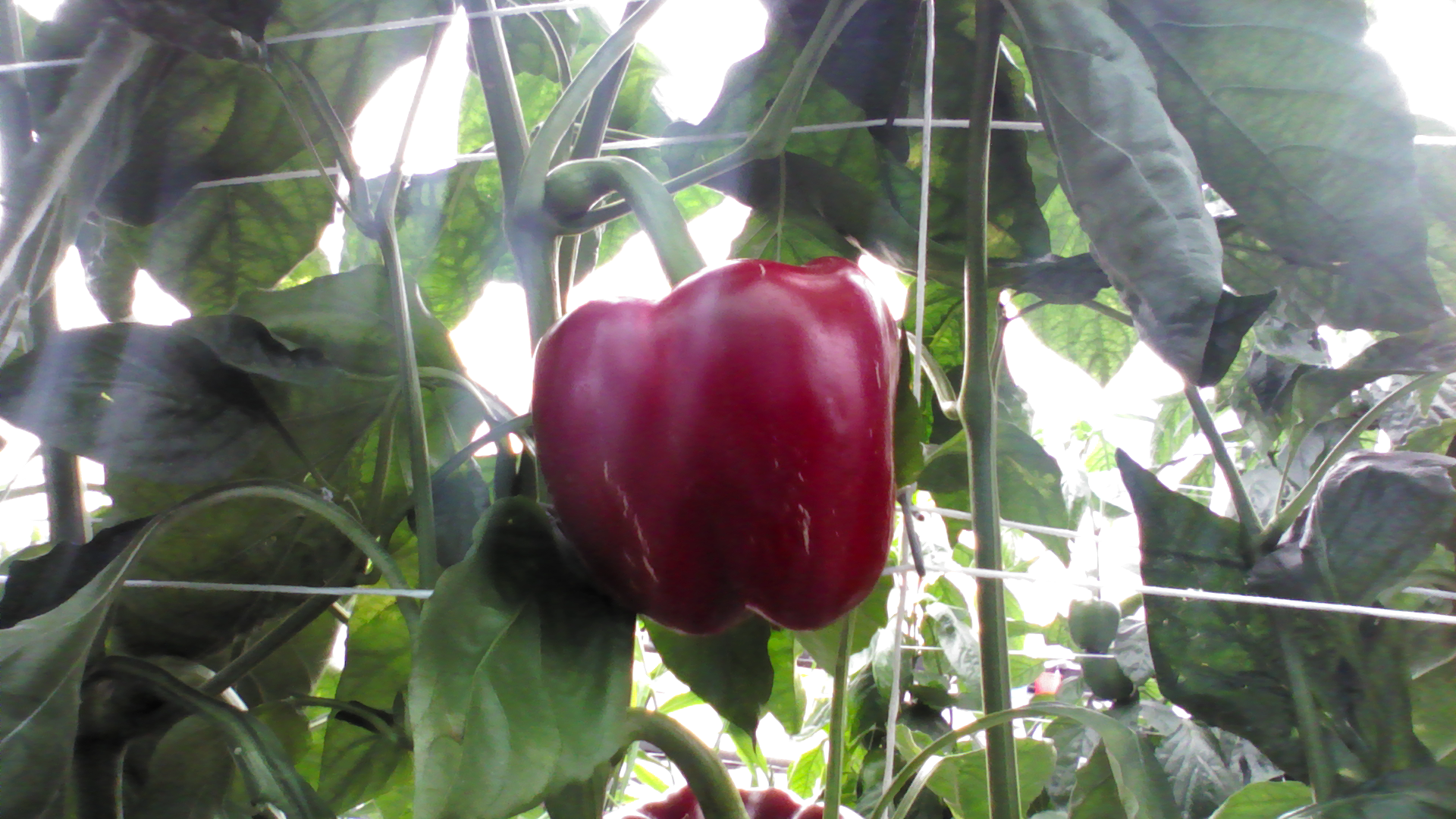}
\includegraphics[width=0.3\textwidth]{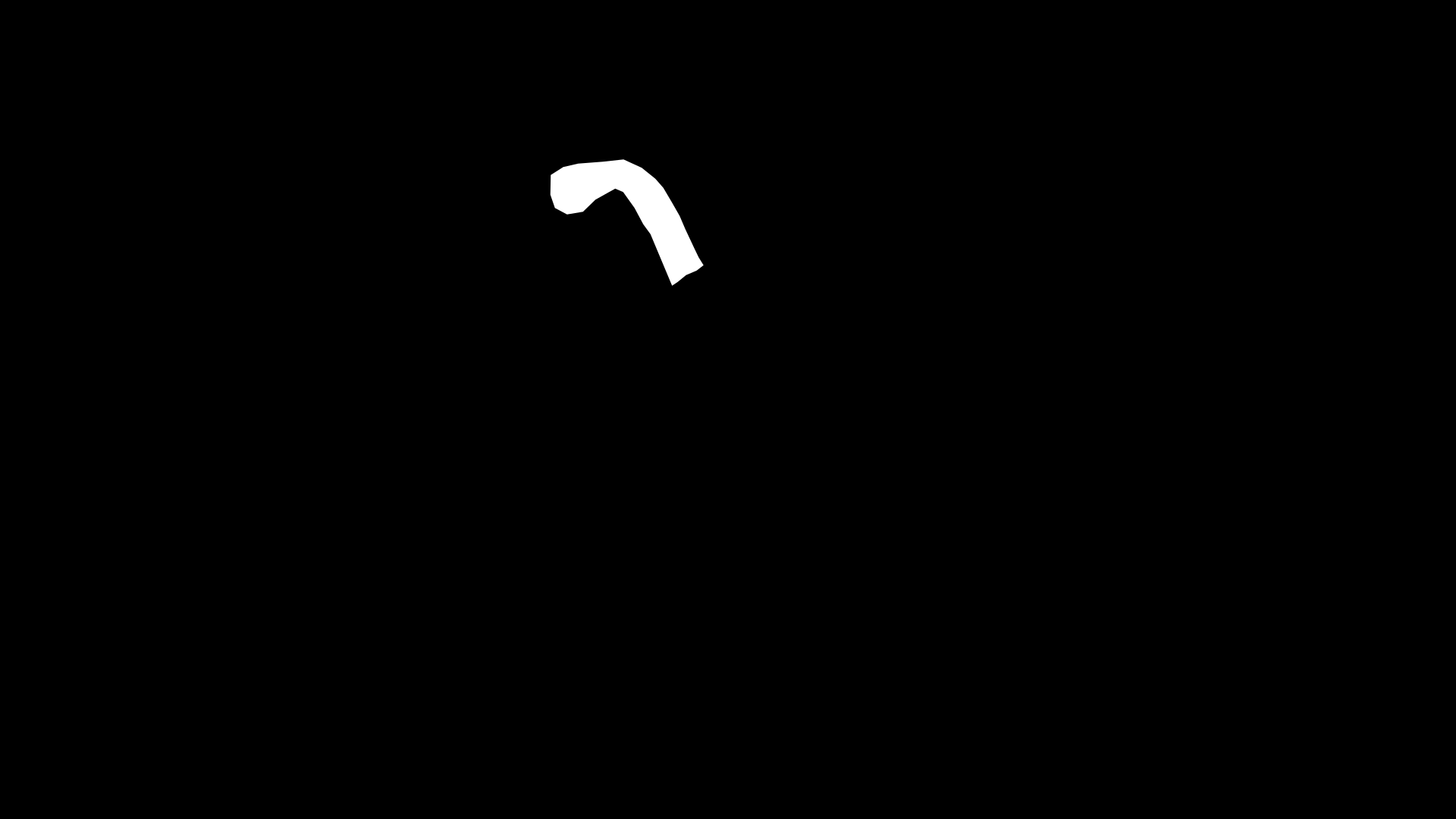}
\includegraphics[width=0.3\textwidth]{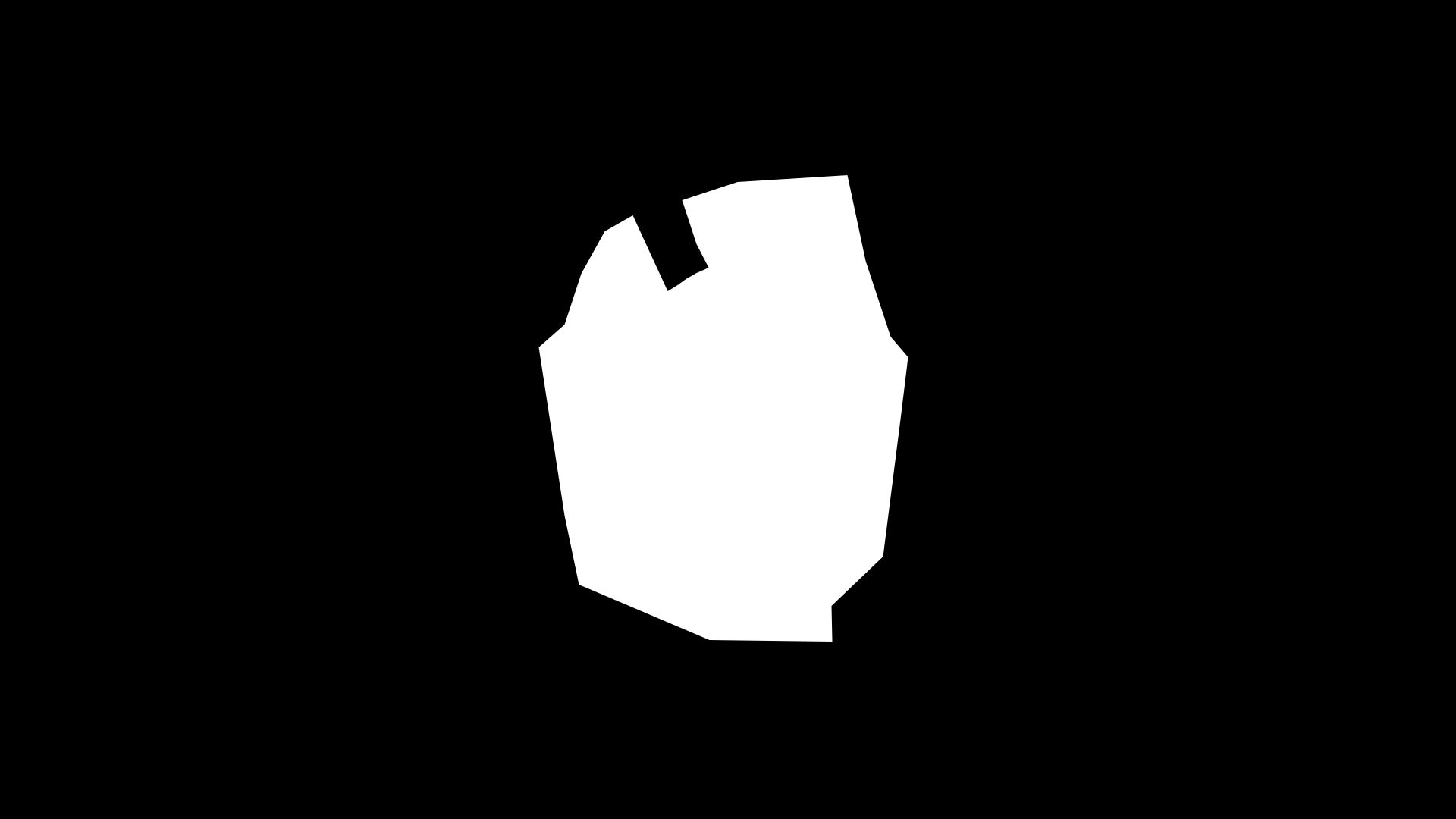}
\end{center}
\vspace{-10pt}
	\caption{Example of manually annotated ground truth for peduncles. The original image is shown in (a) and the annotated peduncle (b) followed by the regions which do not represent the peduncle in (c).}
	\label{fig:cnn_annotations}
\vspace{-5pt}
\end{figure*}

\subsection{Traditional System: colour and geometry features}

Sa et al.~\cite{Sa17_1} described a sweet pepper peduncle detection system that combined colour cues and geometry features.
The colour cues were obtained by transforming the RGB colour image to the more robust hue (H), saturation (S) and value (V) colour space.
The geometry features were obtained by using point feature histograms (PFH)~\cite{rusu2009fast} which make use of the Euclidean distance and normals between the query point $p$ and its $k$ neighbours.
These two sets of features, colour features and geometry features, were then concatenated to form a single feature vector such that each 3D point was described by a $D_{T}=36$ dimensional feature vector consisting of the HSV value ($D_{HSV}=3$) and the PFH ($D_{PFH}=33$).
Hand labelled 3D scenes, see Fig.~\ref{fig:dataset_instances}, were then annotated to provide ground truth data to train a support vector machine (SVM) to classify each point as being either a peduncle (positive class) or not a peduncle (negative class).

\subsection{Deep Learnt System: lightweight agricultural DCNN}

Deep learning~\cite{LeCun15_1}, and in particular deep convolutional neural networks (DCNNs), has opened the potential to jointly learn the most appropriate features and classifier from data.
It is beginning to see greater use within robotic vision and image classification tasks, however, it has not yet received widespread acceptance and use for challenging agricultural robotic tasks.
One of the reasons for this has been the difficulty in training a DCNN from limited data and that many state-of-the-art networks are large, making their deployment on resource limited robotic platforms difficult.
For instance, even the relatively small Inception-v3 model consists of 25M parameters which is small compared to the 180M parameters for VGG~\cite{Szegedy15_1:conference}.
McCool et al.~\cite{McCool17_1} proposed an approach for training deep convolutional neural networks (DCNNs) that allows for the tradeoff between complexity (e.g. memory size and speed) with accuracy while still training an effective model from limited data.
In particular we make use of the \textit{MiniInception} structure, see Figure~\ref{fig:miniincetion_arch}.

We use the approach of McCool et al. to train a lightweight DCNN for efficient peduncle detection.
When training a model like this it is normal to define the positive region, in this case the peduncle, and then consider everything else to be a negative example.
However, due to scene complexity this is not appropriate for our work as some parts of the scene may contain other peduncles, as can be seen in Figure~\ref{fig:cnn_annotations} (a).
As such, for our work we define the positive region, see Figure~\ref{fig:cnn_annotations} (b), as well as the negative region, see Figure~\ref{fig:cnn_annotations} (c).

\begin{figure}
\begin{center}
\includegraphics[width=\columnwidth]{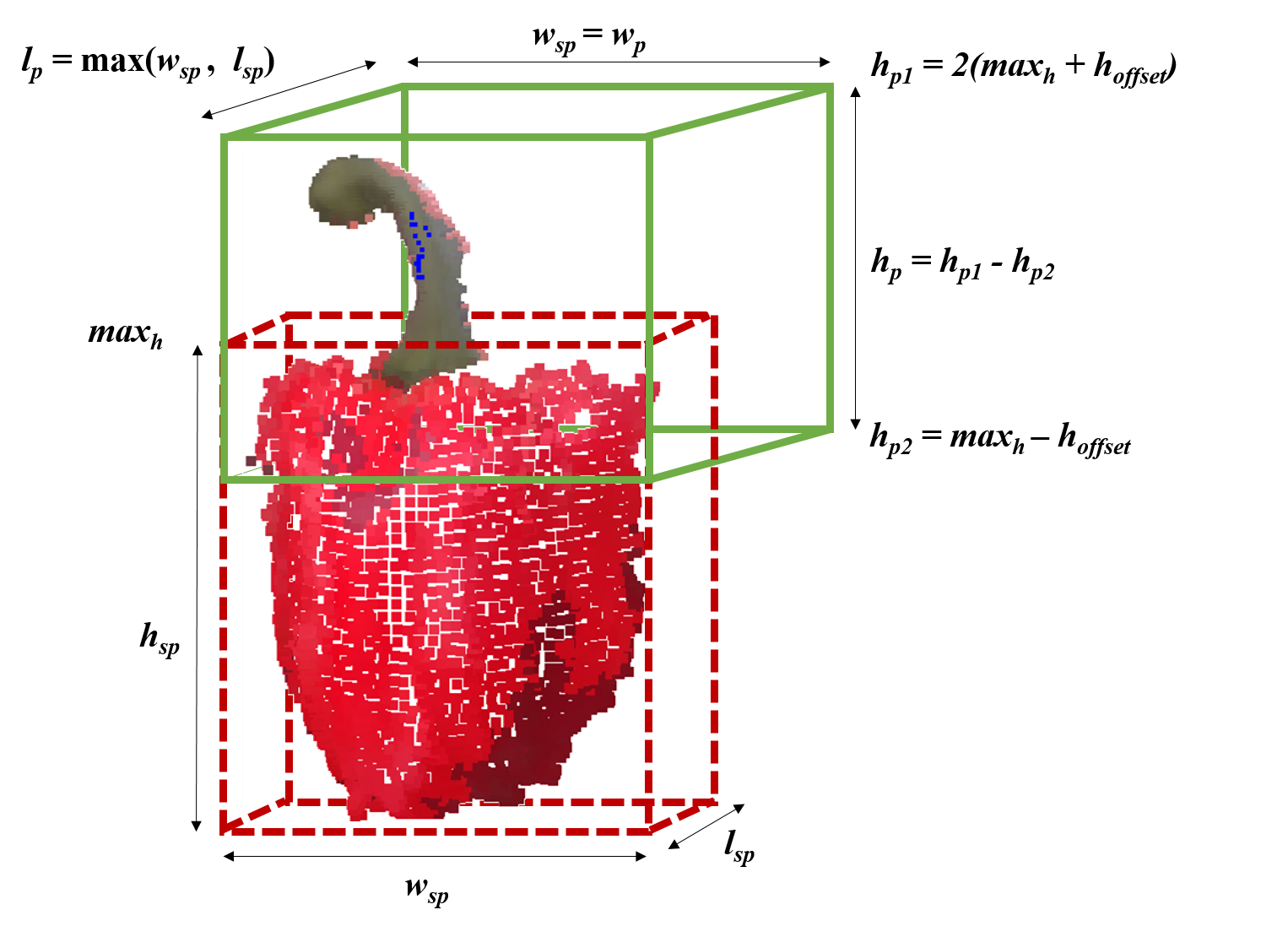}
\end{center}
\vspace{-10pt}
	\caption{Diagram illustrating how a 3D peduncle constraint bounding box is calculated based on the 3D Bounding Box (BB) from the detected sweet pepper. Where $w_{sp}, l_{sp}, h_{sp}$ correspond to the width, length and height of the sweet pepper point cloud and $w_p, l_p, h_p$ corresponds to width, length and height of the calculated peduncle 3D BB. $max_h$ corresponds to the maximum height of the sweet pepper and $h_{offset}$ is a offset parameter for defining the upper and lower heights of the peduncle 3D BB.}
	\label{fig:bbox_diagram}
\vspace{-5pt}
\end{figure}

%
%

\subsection{Region of Interest Selection and 3D Filtering} 

For a deployed system, the task of peduncle detection is usually performed once the crop has been detected.
This means we can employ assumptions based on the structure of the plant to simplify the task.
We make use of two assumptions.
First, we can improve the efficiency of the two algorithms by pre-computing a 2D region of interest (RoI) so that only the region in the image above the detected sweet pepper is considered to contain the peduncle.
Similarly, 3D constraints such that the peduncle cannot be too distant from the detected sweet pepper are enforced using a 3D bounding box before finally declaring the position and pose of the peduncle.
An example of this process is given in Figure~\ref{fig:pipe_line}, this process is applied to both algorithms.

The 2D RoI is defined to contain the region within the image above the detected sweet pepper. Firstly, a bounding box of the detected sweet pepper is computed. The region of interest was then defined to have the same width and height as the sweet pepper bounding box. Lastly, the peduncle 2D bounding box was then shifted up by half the height of the sweet pepper with respect to the centre of the sweet pepper bounding box.


In order to improve the detection of peduncle points a corresponding depth map is used to filter the points using Euclidean constraints.
The filtering steps used are as follows: 
\begin{enumerate}
    \item threshold the classification scores,
    \item project thresholded scores and corresponding depth points to a point cloud,
    \item detect and delete capsicum points using HSV detector,
    \item delete points outside a 3D bounding box and
    \item perform Euclidean clustering on the 3D point cloud and select the largest cluster.
\end{enumerate}

The 3D Bounding Box (BB) used to delete peduncle outliers was computed by using the maximum and minimum euclidean points of the detected sweet pepper points. The definitions of the width, length and height of the BB are defined in \ref{fig:bbox_diagram} and shows that the length, $l_p$ of the peduncle BB is given by the max of either the width, $w_{sp}$, or length, $l_{sp}$ of the sweet pepper BB. The reason for selecting the max of width or length is due to the fact that the depth points of the sweet pepper are measured from one side (view) only and it is assumed that the sweet pepper is mostly symmetrical about this axis. Therefore, the largest measure, width or length, gives the maximum BB of the sweet pepper in those axes. Furthermore, the height $h_p$ of the peduncle BB is defined by the max height of the sweet pepper, $max_h$, and a predefined height offset parameter, $h_{offset}$. For this work we defined the height offset parameter as \unit{50}{mm} which is the average length of a peduncle for the varieties in the field tests.

\begin{figure}[!t]
    \centering
	\begin{subfigure}[b]{0.45\columnwidth}\centering
		\includegraphics[width=0.95\textwidth]{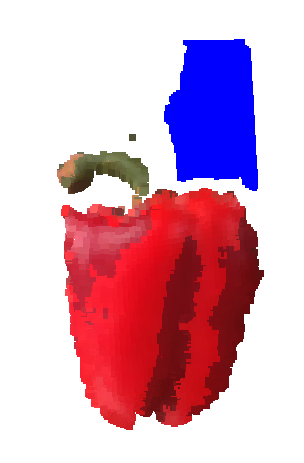}
		\caption{}
	\end{subfigure}
	\begin{subfigure}[b]{0.45\columnwidth}\centering
		\includegraphics[width=0.95\textwidth]{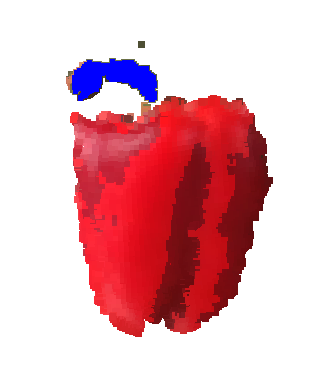}
		\caption{}
	\end{subfigure}	  
	\caption{Example behaviour of the post filtering results where blue represents the classified peduncle points. (a) At low precision (low threshold) based on the assumption of selecting the maximum cluster size, sometimes a cluster that is a separate leaf or other matter is selected. (b) Once the precision becomes higher leaves and other matter are thresholded out and the maximum cluster assumption becomes valid, resulting in the cluster of the peduncle been selected. Therefore in order for a peduncle cluster to be selected a minimum level of precision (threshold) is required.}
    \label{fig:pr_curve_recall_problem}
\end{figure}

These filtering steps require specific parameters, such as model parameters for the HSV detector and minimum, maximum and cluster tolerance for the Euclidean clustering step. The method for detecting sweet peppers points using a prior HSV model and a naive Bayes classifier is described in previous work \cite{lehnert2017autonomous}. The minimum and maximum cluster size was set to 5 points and 25,000 points respectively. The cluster tolerance was set to \unit{3}{mm}.
%
	
%
	

\section{Experimental Results}\label{sec:results}

We present results for the two algorithms, \textit{PFH-SVM} and \textit{MiniInception}, executed on Harvey\footnote{A small form GeForce 1070 was used for inference of the \textit{MiniInception} model.}.
The system was deployed in a glasshouse facility in Cleveland (Australia) and consisted of two cultivar \textit{Ducati} and \textit{Mercuno}.
To train the \textit{MiniInception} approach 41 annotated images were used. These images came from two sites, 20 images were obtained from the same site in Cleveland several weeks prior to deploying the robot which included a different set of crop on the plant and 21 were obtained from another site in Giru, North Queensland.

\subsection{Segmentation Performance}

\begin{figure}[!t]
    \centering
    \includegraphics[trim={3cm 0 3cm 0},clip, width=\columnwidth]{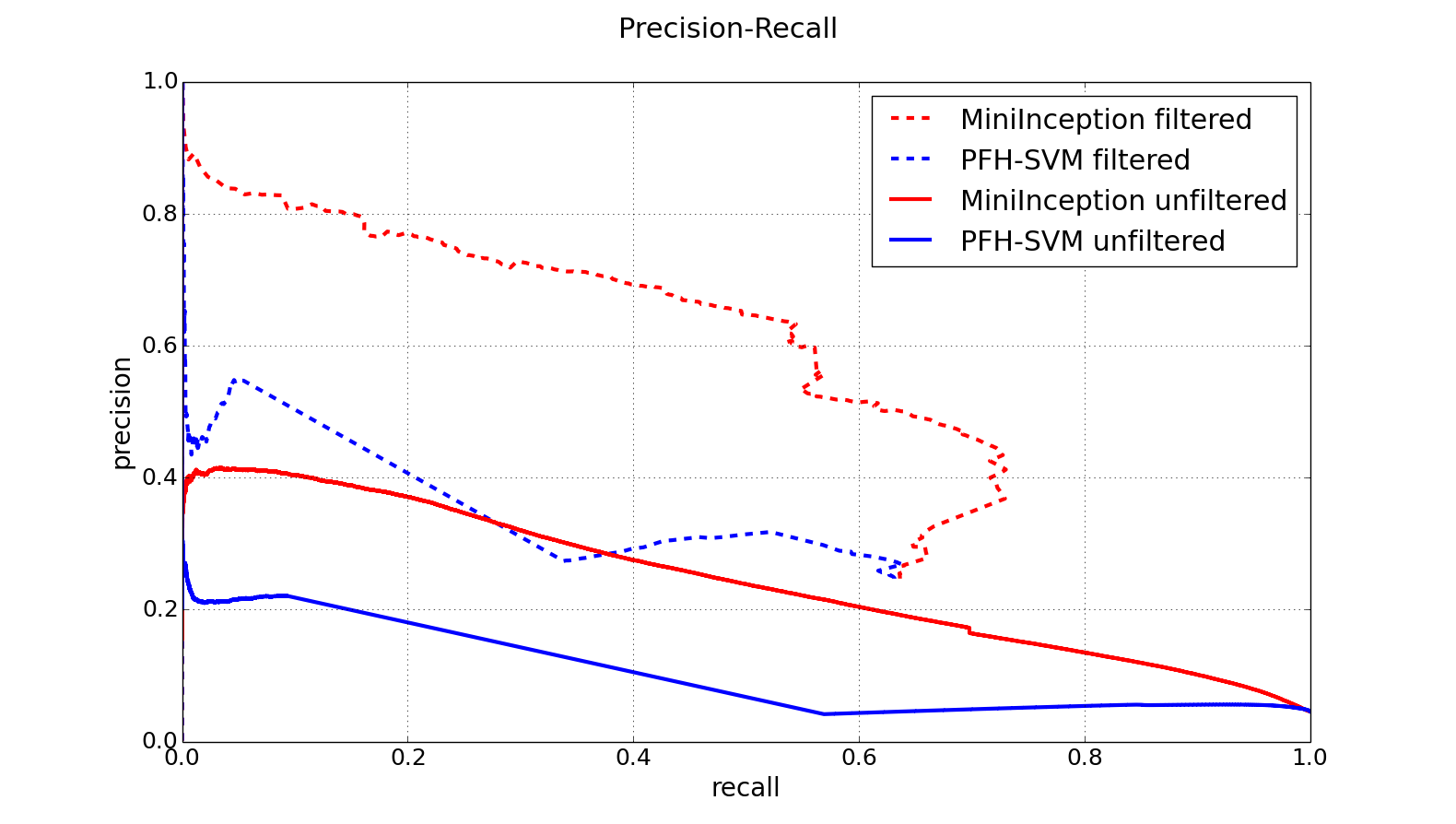}
    \caption{Precision recall results for the \textit{MiniInception} (DeepNet) and \textit{PFH-SVM} (PFH) algorithms before and after the filtering step.}
    \label{fig:pr_dcnn_vs_trad}
\end{figure}

The performance of the two algorithms, \textit{PFH-SVM} and \textit{MiniInception}, is summarised in Figure~\ref{fig:pr_dcnn_vs_trad}.
It can be seen that the performance of the \textit{MiniInception} model is consistently superior to that of the \textit{PFH-SVM} approach.
However, both approaches have relatively low performance with $F_1$ scores of 0.313 and 0.132 for the \textit{MiniInception} and \textit{PFH-SVM} systems respectively.
On average the execution time of the two algorithms is similar with the \textit{MiniInception} approach executing an average of 1704 points per second while the \textit{PFH-SVM} approach executes at an average of 1248 points per second. This measurement is reported as the two methods receive a different number of points (3D points vs 2D pixels) for the same data.

Introducing the filtering step provides a considerable improvement in performance for both algorithms.
The $F_1$ for the \textit{MiniInception} and \textit{PFH-SVM} systems improve to 0.564 and 0.302 respectively.
For both algorithms, introducing the filtering step leads to odd behaviour in the precision-recall curve.
This is expected because we are altering the threshold on an algorithm, either \textit{MiniInception} or \textit{PFH-SVM}, whose performance is dependent an another step which greatly impacts its performance.
An example of this is illustrated in Figure~\ref{fig:pr_curve_recall_problem} where introducing the filtering step at different thresholds leads to different points of the 3D point cloud being considered as peduncles, and other points being suppressed.

\subsection{Qualitative Results}

\begin{figure}[!t]
    \centering
    \includegraphics[trim={1cm 0 2cm 0},clip, width=\columnwidth]{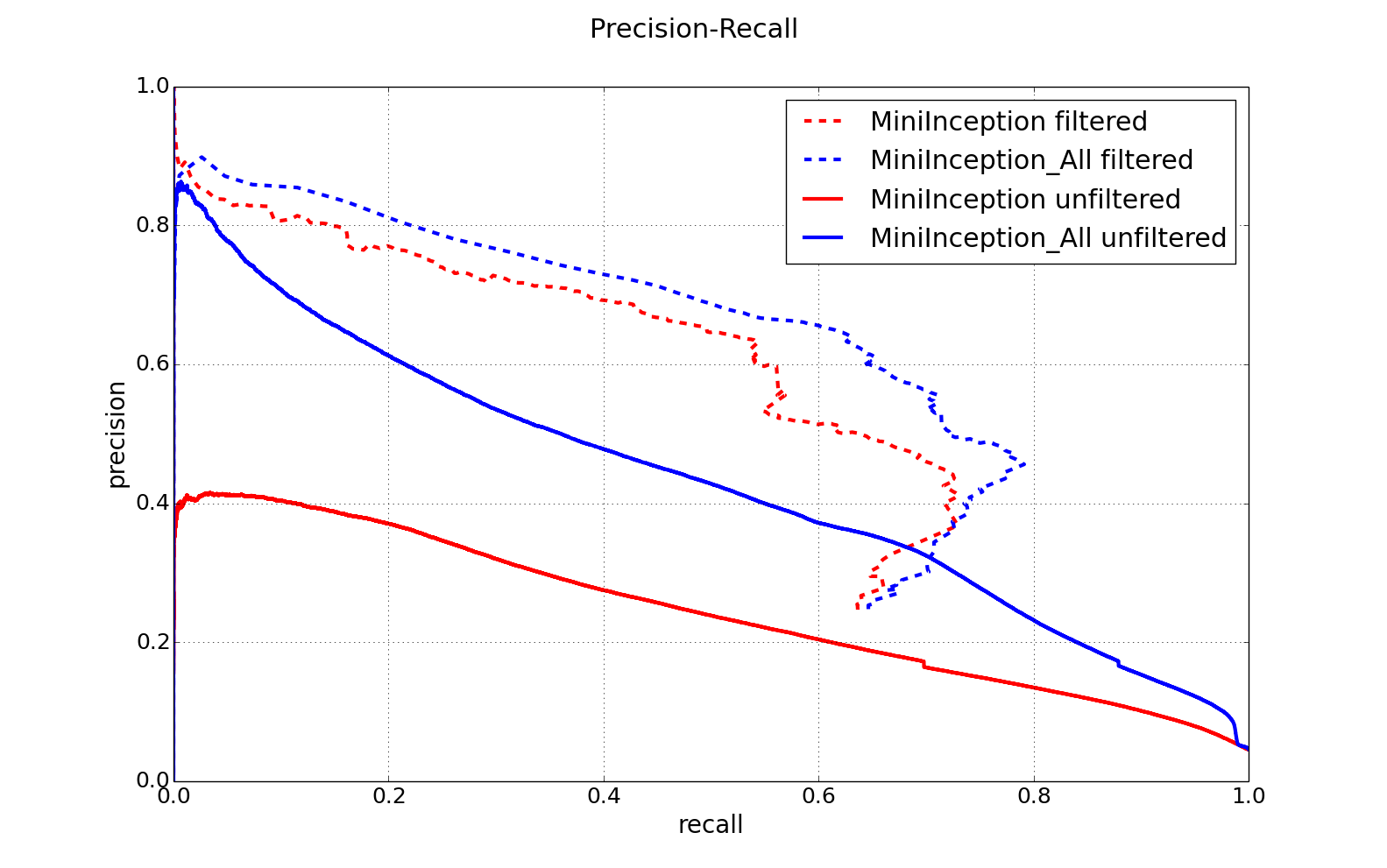}
    \caption{Precision recall results for the \textit{MiniInception}, before and after the filtering step, using different amount of training data.}
    \label{fig:extended_pr_dcnn_vs_trad}
\end{figure}

\begin{figure}[tb]
	\centering
	\begin{subfigure}[b]{0.49\columnwidth}\centering
		\includegraphics[width=0.95\textwidth]{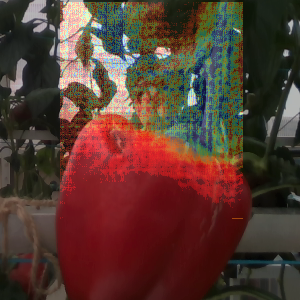}	
	\end{subfigure}
	\begin{subfigure}[b]{0.49\columnwidth}\centering
		\includegraphics[width=0.95\textwidth]{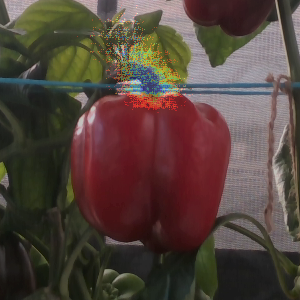} 
	\end{subfigure}	
	\vspace{1pt}
	
	\begin{subfigure}[b]{0.49\columnwidth}\centering
		\includegraphics[width=0.95\textwidth]{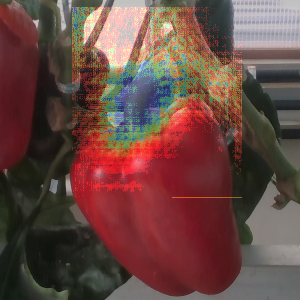} 	
	\end{subfigure}
	\begin{subfigure}[b]{0.49\columnwidth}\centering
		\includegraphics[width=0.95\textwidth]{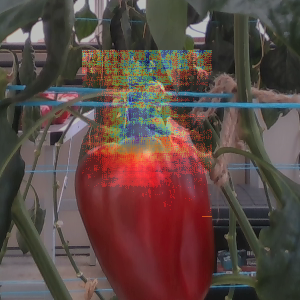}
	\end{subfigure}	
	\vspace{1pt}

	\begin{subfigure}[b]{0.49\columnwidth}\centering
		\includegraphics[width=0.95\textwidth]{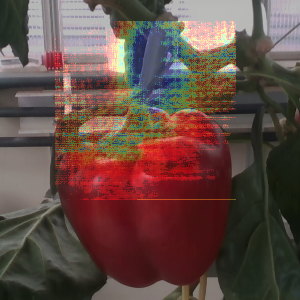}
	\end{subfigure}
	\begin{subfigure}[b]{0.49\columnwidth}\centering
		\includegraphics[width=0.95\textwidth]{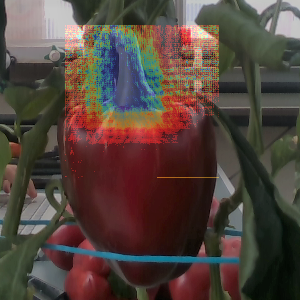}
	\end{subfigure}
	\caption{Example peduncle detection responses from CNN. It can be seen that some detections are made not only on the peduncle but on the stem of the plant as well. Most stem detections can be filtered out using further Euclidean clustering and constraints on the resulting point cloud of all detected points}
	\label{fig:CNN_image_results}
\end{figure}

Qualitative results for the \textit{MiniInception} segmentation algorithm are presented in Figure~\ref{fig:CNN_image_results}.
From these results it can be seen that the deep network approach provides consistent results across multiple poses.
Also, it can be seen that the regions with high scores surround the peduncle region.
We believe this, in part, explains the poor precision-recall curve for the \textit{MiniInception} algorithm as these points will be considered as false positives and greatly reduce the precision value.
This is despite their proximity to the peduncle.
This also explains the considerable gain achieved by introducing the filtering step as many of these points will correspond to background regions and be discarded.

\begin{figure*}[tb]
	\centering
	\begin{subfigure}[b]{0.19\textwidth}\centering
		\includegraphics[trim={4cm 1cm 4cm 1cm},width=0.95\textwidth]{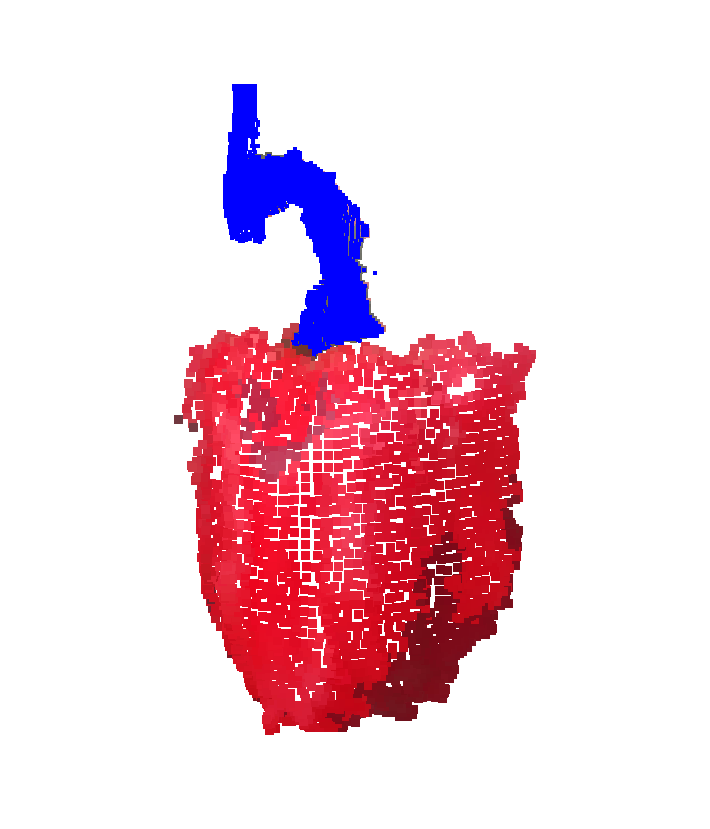}	
	\end{subfigure}
	\begin{subfigure}[b]{0.19\textwidth}\centering
		\includegraphics[trim={4cm 1cm 4cm 1cm},width=0.95\textwidth]{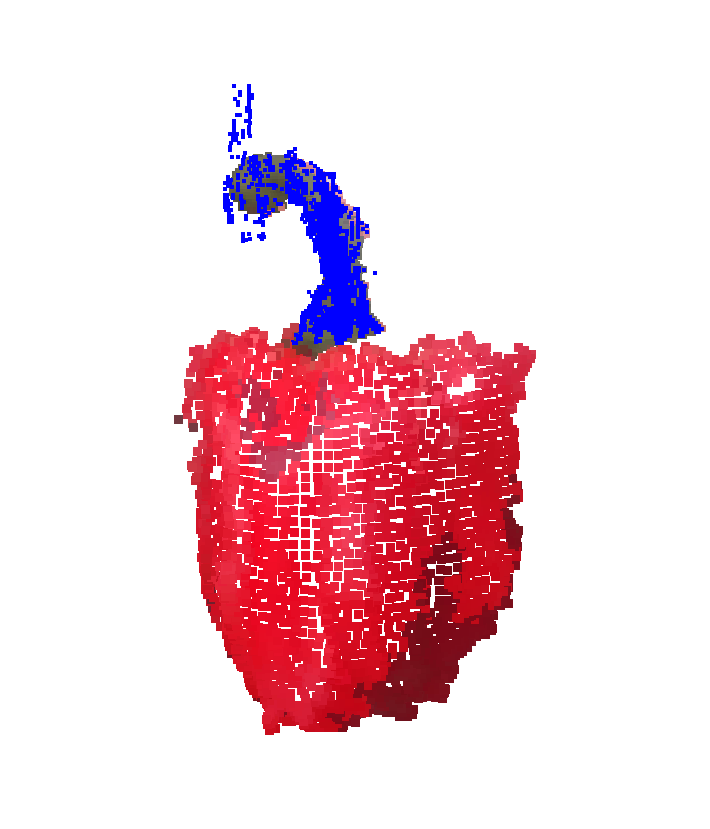} 
	\end{subfigure}	
	\begin{subfigure}[b]{0.19\textwidth}\centering
		\includegraphics[trim={4cm 1cm 4cm 1cm},width=0.95\textwidth]{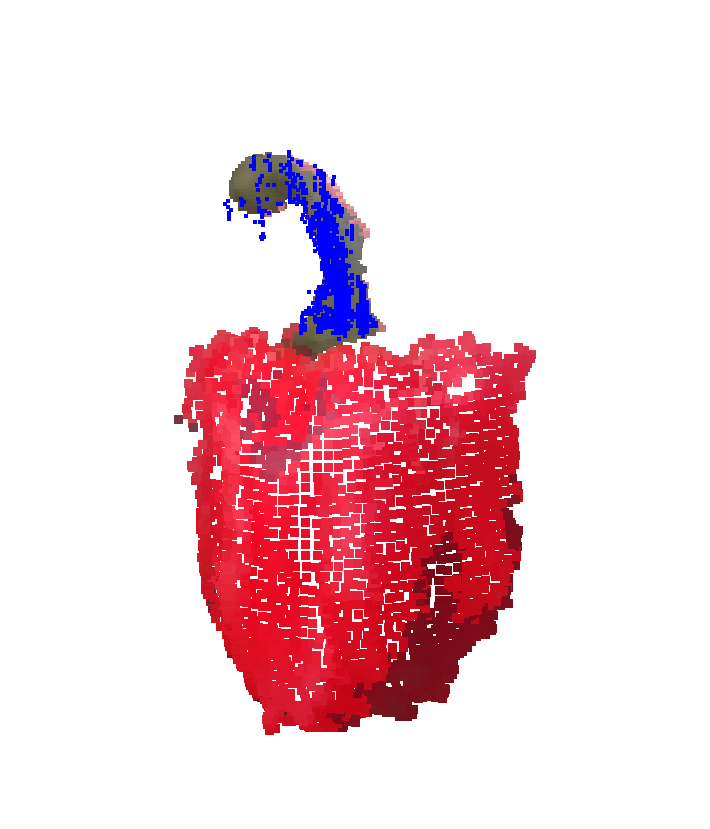} 	
	\end{subfigure}
	\begin{subfigure}[b]{0.19\textwidth}\centering
		\includegraphics[trim={4cm 1cm 4cm 1cm},width=0.95\textwidth]{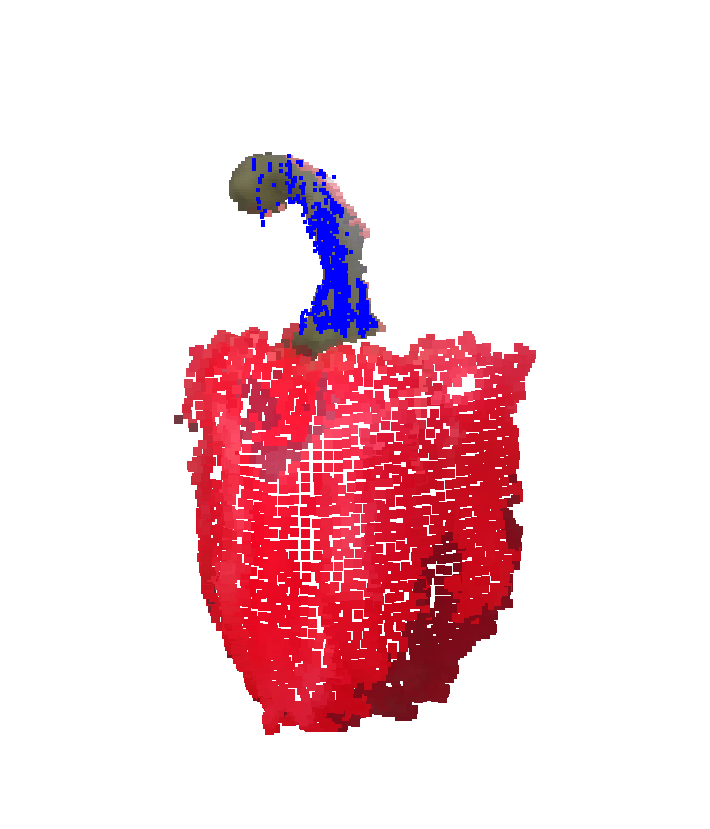}
	\end{subfigure}	
	\begin{subfigure}[b]{0.19\textwidth}\centering
		\includegraphics[trim={4cm 1cm 4cm 1cm},width=0.95\textwidth]{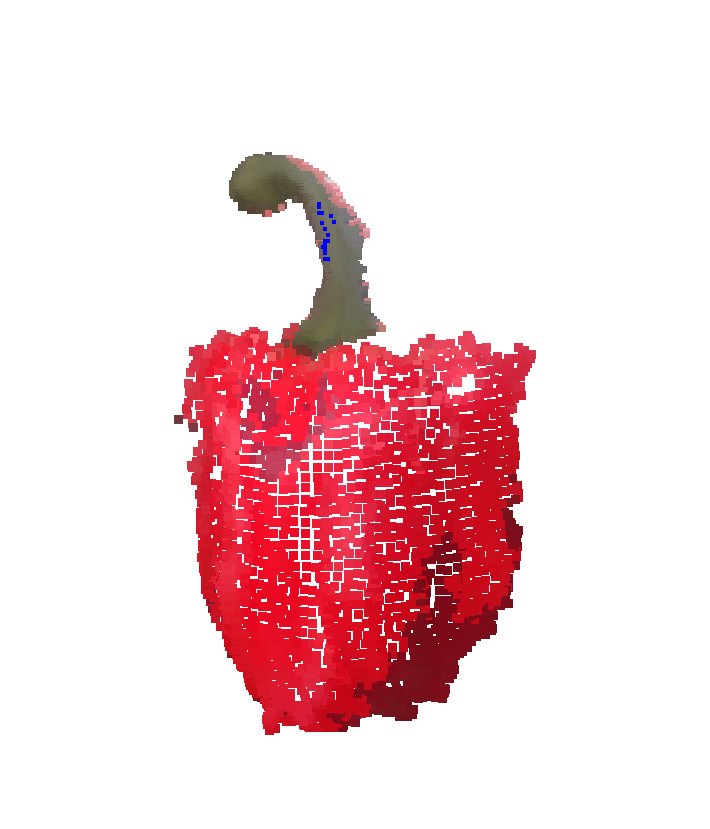}
	\end{subfigure}
	\caption{Example peduncle detection after filtering from CNN responses with varying threshold values.
	The threshold value increases from left to right. The detected peduncle points are highlighted in blue.}
	\label{fig:CNN_and_pfh_filter_results}
\end{figure*}

In Figure~\ref{fig:CNN_and_pfh_filter_results} we present example results of the entire procedure for the \textit{MiniInception} algorithm, with filtering, at varying thresholds.
It can be seen that as the threshold is increased the erroneous points, such as those belonging to the stem, are removed.
Even at higher threshold values a large number of points on the peduncle of the fruit are chosen.

\subsection{Extended Training Data}

One of the advantages of the \textit{MiniInception} approach is that it is much easier to annotate training than the \textit{PFH-SVM} approach.
To determine if this can be beneficial we extended the training set of \textit{MiniInception} with an extra 33 images\footnote{These images came from the Cleveland test site.}, to a total of 74 annotated images.
This system is referred to as \textit{MiniInception-Extended}.

In Figure~\ref{fig:extended_pr_dcnn_vs_trad} it can be seen that the \textit{MiniInception} approach benefits considerably by increasing the training set size.
The $F_1$ improves from 0.313 for \textit{MiniInception} to 0.452 for \textit{MiniInception-Extended}, a relative improvement of 31\%.
Including the filtering step again provides a boost in performance leading an $F_1$ of 0.631.
This is a relative improvement of 10.8\% and demonstrates one of the key potential advantages of this deep learning approach that it benefits from increasing the training set size and annotating the training data is relatively easy as it required the labelling of a 2D image rather than a 3D point cloud (as is the case for the \textit{PFH-SVM} approach).

\section{Summary and Conclusion}\label{sec:conclusion}

We have presented a comparative study of two peduncle detection systems, \textit{PFH-SVM} and \textit{MiniInception}, deployed on a robotic platform, Harvey.
Trials on two cultivar found that the deep learning approach, \textit{MiniInception}, provided superior performance with an $F_1$ of 0.564 with filtering.
In addition to superior performance, \textit{MiniInception} had the added advantage that it has a simpler annotation process as it requires 2D images to be annotated.
By comparison \textit{PFH-SVM} requires the annotation of 3D point clouds.

Further experiments demonstrated that with more data the \textit{MiniInception} model is able to achieve considerably higher accuracies.
Extending the training data by 80\% provided a relative improvement in performance of the system of 10.8\%, with filtering.
This is a critical advantage given that is much easier to annotate 2D images than 3D point clouds.

Future work will evaluate the impact this approach has on the final harvesting rate of a system such as Harvey.

\section*{Acknowledgements}
The authors would like to thank Inkyu Sa for his key contribution in the peduncle detection algorithm and assistance deploying his approach to Harvey. 
Thanks also goes to Raymond Russell and Jason Kulk of QUT Strategic Investment in Farm Robotics Program for their key contributions to the design and fabrication of the harvesting platform. 
We would also like to acknowledge Elio Jovicich and Heidi Wiggenhauser from Queensland Department of Agriculture and Fisheries for their advice and support during the field trips. 

\balance 

\bibliographystyle{ieeetr}
\bibliography{bibs/IROS2016}

\end{document}